\documentclass[conference]{IEEEtran}

\makeatletter
\def\@copyrightspace{\relax}
\makeatother

\usepackage{balance}  
\usepackage{graphics} 
\usepackage{times}    
\usepackage{url}      
\usepackage{float}
\usepackage[space]{cite}

\makeatletter
\def\url@leostyle{%
  \@ifundefined{selectfont}{\def\UrlFont{\sf}}{\def\UrlFont{\small\bf\ttfamily}}}
\makeatother
\usepackage[table,xcdraw]{xcolor}
\usepackage{multirow}
\usepackage{cleveref}

\usepackage{graphicx}
\usepackage{subcaption}

\begin{document}

\title{Assessing the Impact of Attention and Self-Attention Mechanisms on the Classification of Skin Lesions}

\author{\IEEEauthorblockN{Rafael Pedro}
\IEEEauthorblockA{\textit{Instituto Superior Técnico} \\
Lisbon, Portugal \\
rafael.pedro@tecnico.ulisboa.pt}
\and
\IEEEauthorblockN{Arlindo L. Oliveira}
\IEEEauthorblockA{\textit{INESC-ID / Instituto Superior Técnico} \\
Lisbon, Portugal \\
arlindo.oliveira@tecnico.ulisboa.pt}
}

\maketitle
\begin{abstract}

Attention mechanisms have raised significant interest in the research community, since they promise significant improvements in the performance of neural network architectures. However, in any specific problem, we still lack a principled way to choose specific mechanisms and hyper-parameters that lead to guaranteed improvements. More recently, self-attention has been proposed and widely used in transformer-like architectures, leading to significant breakthroughs in some applications. In this work we focus on two forms of attention mechanisms: attention modules and self-attention. Attention modules are used to reweight the features of each layer input tensor. Different modules have different ways to perform this reweighting in fully connected or convolutional layers. The attention models studied are completely modular and in this work they will be used with the popular ResNet architecture. Self-Attention, originally proposed in the area of Natural Language Processing makes it possible to relate all the items in an input sequence. Self-Attention is becoming increasingly popular in Computer Vision, where it is sometimes combined with convolutional layers, although some recent architectures do away entirely with convolutions. In this work, we study and perform an objective comparison of a number of different attention mechanisms in a specific computer vision task, the classification of samples in the widely used Skin Cancer MNIST dataset.
The results show that attention modules do sometimes improve the performance of convolutional neural network architectures, but also that this improvement, although noticeable and statistically significant, is not consistent in different settings. The results obtained with self-attention mechanisms, on the other hand, show consistent and significant improvements, leading to the best results even in architectures with a reduced number of parameters.

\end{abstract}


\section{Introduction}

Attention refers to the cognitive process of selectively concentrating on some stimuli while ignoring others. Humans tend to focus their attention on certain parts of the visual space, in order to acquire information, instead of seeing the image as a whole. We focus on few things at a time and then combine the information retrieved from them. The use of attention in human vision is a way to save brain power, making it easier to analyze complex scenes. Mechanisms of attention are already used in many application domains. Two of them, Natural Language Processing (NLP) \cite{aiayn,ltpa} and Computer Vision (CV) tasks \cite{squeeze,cbam,ta,aa,eca,vit,convit}, are especially relevant.

When reading a book, we tend to focus on a word at a time, and with the contextual relevance of the previous words, we can make sense of a phrase. Our brain uses attention when reading, so applying similar strategies to neural networks for NLP tasks would only make sense.

Vision tasks also make use of attention. If we look at a big family photo, our brain will have a blurry perception of the image as a whole. To fully understand who is in the picture, we will pay attention to each face, look at clothes individually, and see if we can identify where the photo was taken. In this way, we can have a better understanding of what is happening. The intuition behind attention mechanisms in computer vision tasks is precisely this.

Attention improves the representation of stimuli. By focusing on essential features and suppressing unnecessary ones, our brain can increase representation power \cite{cbam}, therefore reducing the noise it has to process. Throughout this paper, the focus will be on attention mechanisms applied to computer vision tasks.

There is no single precise definition of "attention" for neural networks, but the central idea behind attention mechanisms is that they enable the network to selectively pay attention to specific features, or to combinations of features, that the architecture would not allow if attention mechanisms were not used. In most cases, attention mechanisms correspond to neural network layers that combine information from previous layers \cite{detr}, enabling deeper layers to use this aggregate information. 

Attention and, especially, self-attention mechanisms became popular in NLP tasks, but their use in Computer Vision has been increasing for several years. Although attention mechanisms in computer vision are relatively recent, many different approaches have already been proposed to date.

The basis of Convolutional Neural Network (CNN) architectures is the convolution operation. A CNN performs a convolution with a local receptive field, defined by the kernel size, making it hard for a specific layer of the network to explore information outside of this local receptive field. This creates a limitation in CNNs, as a single layer can only combine inputs (pixels or voxels) that are near each other. The absence of global contextual information is a focus of study in the attention mechanisms mentioned in the upcoming sections.

The primary backbone of the Transformer architecture is self-attention. Self-attention is a mechanism that relates the information in different positions of a sequence to build a representation of the sequence \cite{aiayn}, creating a sequence-to-sequence mapping. Each layer of a Transformer receives a sequence as an input and outputs another sequence. Transformers are a different type of neural network architecture, composed of repeating blocks of self-attention and fully-connected feed-forward layers. In this work, we only make use of the encoder. 

For the purposes of this work, attention mechanisms can therefore be classified in two different classes: attention modules and self-attention. 
\begin{itemize}
\item Attention modules are blocks inserted into fully connected or, most commonly, convolutional layers, that change the way the layers process the input of previous layers. Four different attention modules were studied. All of them were tested with the different variants of the ResNet \cite{resnet} architecture.
\item  Self-attention mechanisms are used inside Transformer blocks. These blocks are then stacked to create a network. Self-attention can be combined with fully connected layers or with convolutional neural network layers. 
\end{itemize}

\section{Attention in Neural Networks}

 In this paper, we study only attention mechanisms that influence the training process \cite{ltpa}. We do not cover other attention-related methods, such as GradCAM \cite{gradcam}, which produces visual explanations of CNNs, making them more transparent and explainable. However, we will use GradCam to analyse the behavior of networks, in \Cref{sec:att_results}.

Attention mechanisms can be divided into three major classes: hard attention, soft attention and self-attention, the last two being the ones tested in this work.

\subsection{Hard Attention}

Hard attention can be viewed as an on-off switch that determines whether or not the network should pay attention to a specific region. Hard attention usually uses a crop of a specific zone of the input image \cite{ltpa}. However, this method is non-differentiable because discrete variables are used to describe the region of attention. 

Therefore, these architectures need to be trained with a reinforcement learning strategy. One relevant article that makes use of hard attention considers the attention problem as a sequential decision process with an agent, trained using reinforcement learning, that interacts with the environment \cite{rmova}. 

\subsection{Soft Attention \label{sec:soft}}

Soft attention uses functions that vary smoothly over their domain and, therefore, are  differentiable. Soft attention modules are trainable along with the network, using standard back-propagation or any other gradient descent method \cite{ltpa}. 

In computer vision, soft attention is usually preferable to hard attention since it is trainable with the network, and non-rigid attention maps (crops) are less suited to natural shaped objects seen in real-world images \cite{hie}. In the scope of soft attention, two main variants were studied: methods that use only channel attention and methods that use channel and spatial attention \cite{aootamicv}.

\subsubsection{Channel Attention}

When a convolutional layer processes an image, new feature maps are generated by each successive layer. Two dimensional color images initially have three different channels (e.g., red, green and blue, in the RGB system). After being processed by the kernels in the first convolutional layer, a number of new channels are generated with additional information. The original channels may, or may not, be made available to upper layers.

Channel attention mechanisms weight the feature channels according to the importance they hold \cite{aootamicv}. A commonly used proposal, the Squeeze and Excitation (SE) attention mechanism \cite{squeeze}, uses a module that performs a squeeze operation, which compresses the spatial dimensions into a real number, \cite{aootamicv}, and after the excitation operation generates the attention weights. These weights are then used to reweight the original input tensor. 

In this work, we experimented with the Squeeze and Excitation (SE) mechanism and the Efficient Channel Attention (ECA) \cite{eca}. These modules reweight the channel part of the tensor. ECA appears as an improvement to SE by being a channel attention module with fewer parameters. ECA avoids the use of dimensionality reduction by employing a one-dimensional convolution. This convolution has a kernel size that changes according to the number of channels in the input feature map.

\begin{figure}[htbp!]
	\centering	        
    \includegraphics[width=0.5\textwidth]{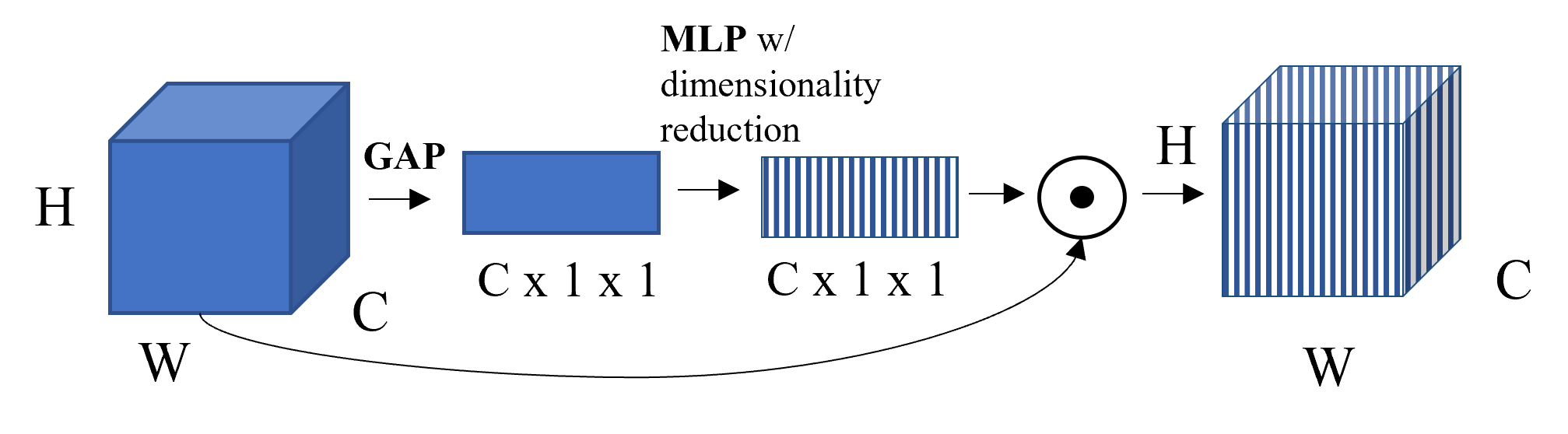}
	\caption{Architecture of the SE attention module. $\odot$ represents element-wise multiplication and GAP stands for Global-Average Pooling.}
	\label{fig:se}
\end{figure}

\begin{figure}[htbp!]
	\centering	        
	\includegraphics[width=0.5\textwidth]{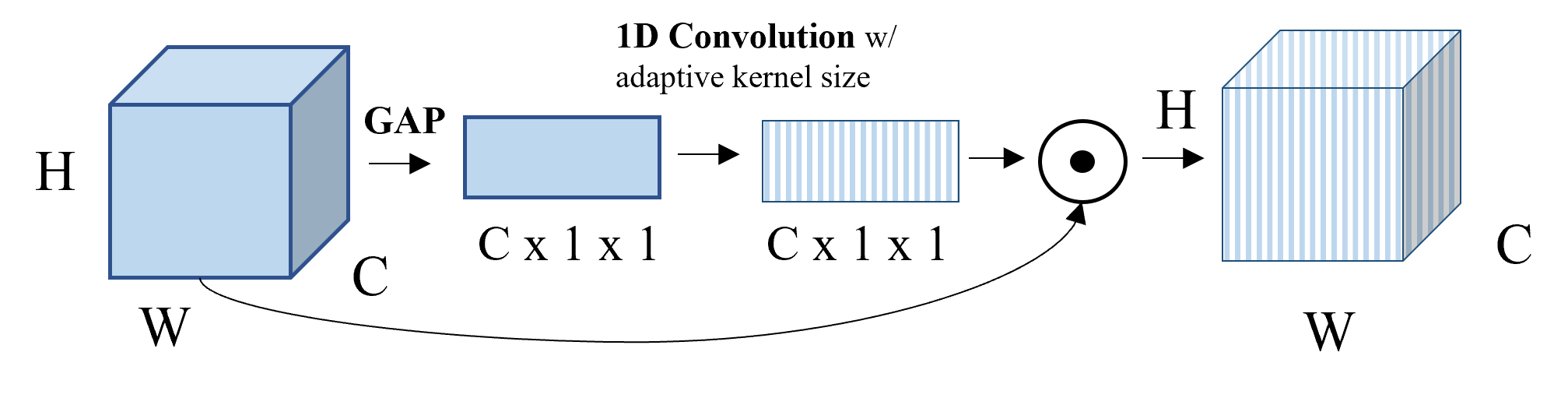}
	\caption{Architecture of the ECA attention module. The main change from SE is the use of a one-dimensional convolution with adaptive kernel size. $\odot$ represents element-wise multiplication and GAP stands for Global-Average Pooling.}
	\label{fig:eca}
\end{figure}

\subsubsection{Channel and Spatial Attention}

After the success of Squeeze and Excitation (SE) \cite{squeeze}, other attention modules have been proposed. The channel and spatial attention modules introduce a new spatial sub-module that processes information from the tensor spatial dimension. Besides reweighting the channel space, these modules usually use convolutional layers to also reweight the spatial part of the input tensor, giving more weight to specific parts of the input image.

The Convolutional Block Attention Module (CBAM) \cite{cbam} and the Triplet Attention (TA) \cite{ta} use extra sub-modules which look spatially to the tensor, and are examples of channel and spatial attention modules. The reweighting performed by these models applies to both the channel and the spatial part of the tensor. These modules use convolutions to create the spatially reweighted tensor. TA was proposed as an improvement to CBAM with a similar architecture and negligible parameters overhead. 

SE and CBAM have been influential in attention mechanisms studies and they have been used in several applications. Nonetheless, ECA and TA came as improvements to the previous modules so they are also tested and studied in this work.

\begin{figure}[htbp!]
	\centering	        
    \includegraphics[width=0.5\textwidth]{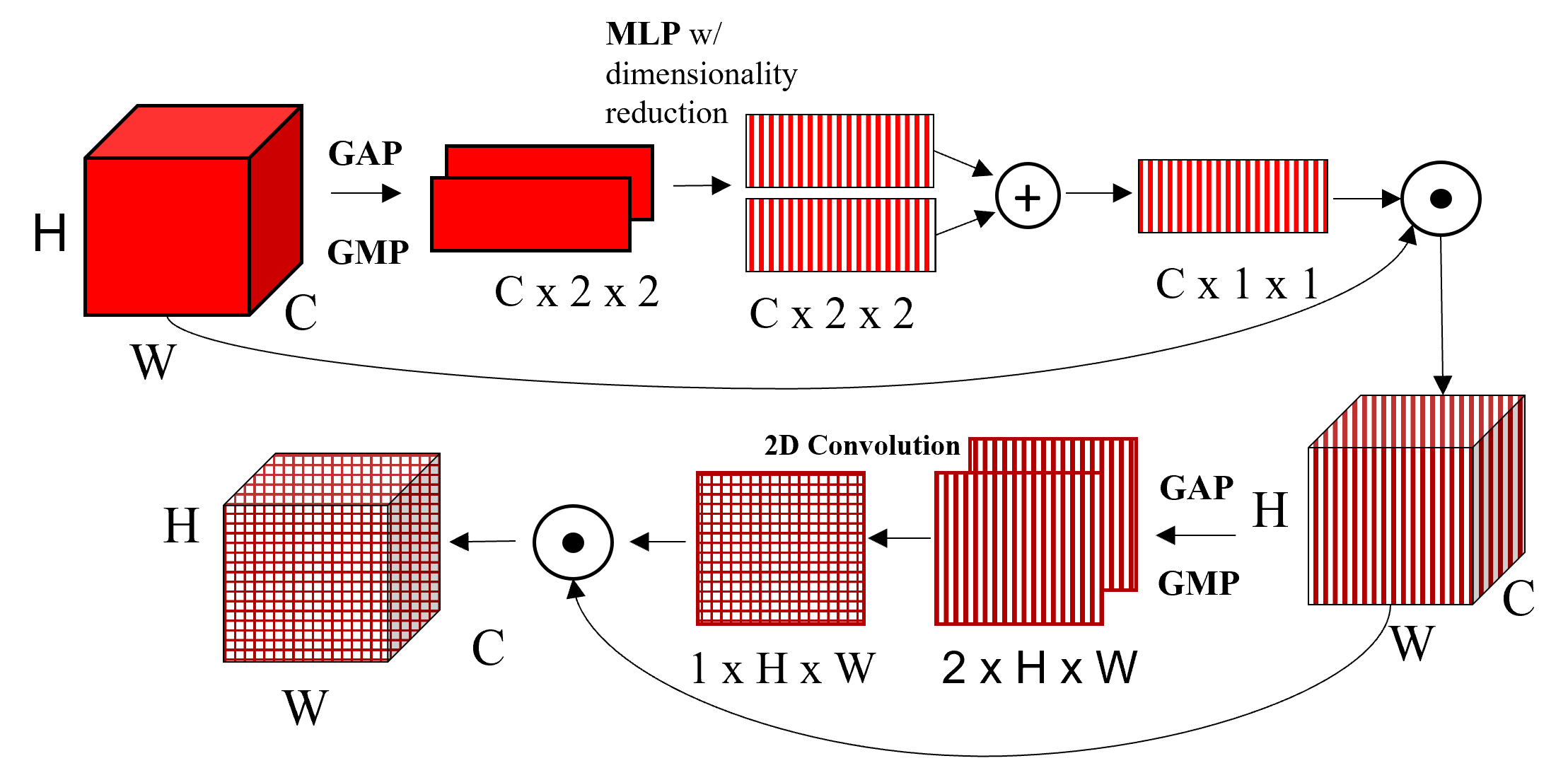}
	\caption{Architecture of the CBAM attention module, divided in the channel attention sub-module and the spatial attention sub-module.  $\odot$ represents element-wise multiplication and $\oplus$ element-wise summation, while GAP stands for Global-Average Pooling and GMP for Global-Max Pooling.}
	\label{fig:cbam}
\end{figure}

\begin{figure}[htbp!]
	\centering       
	\includegraphics[width=0.5\textwidth]{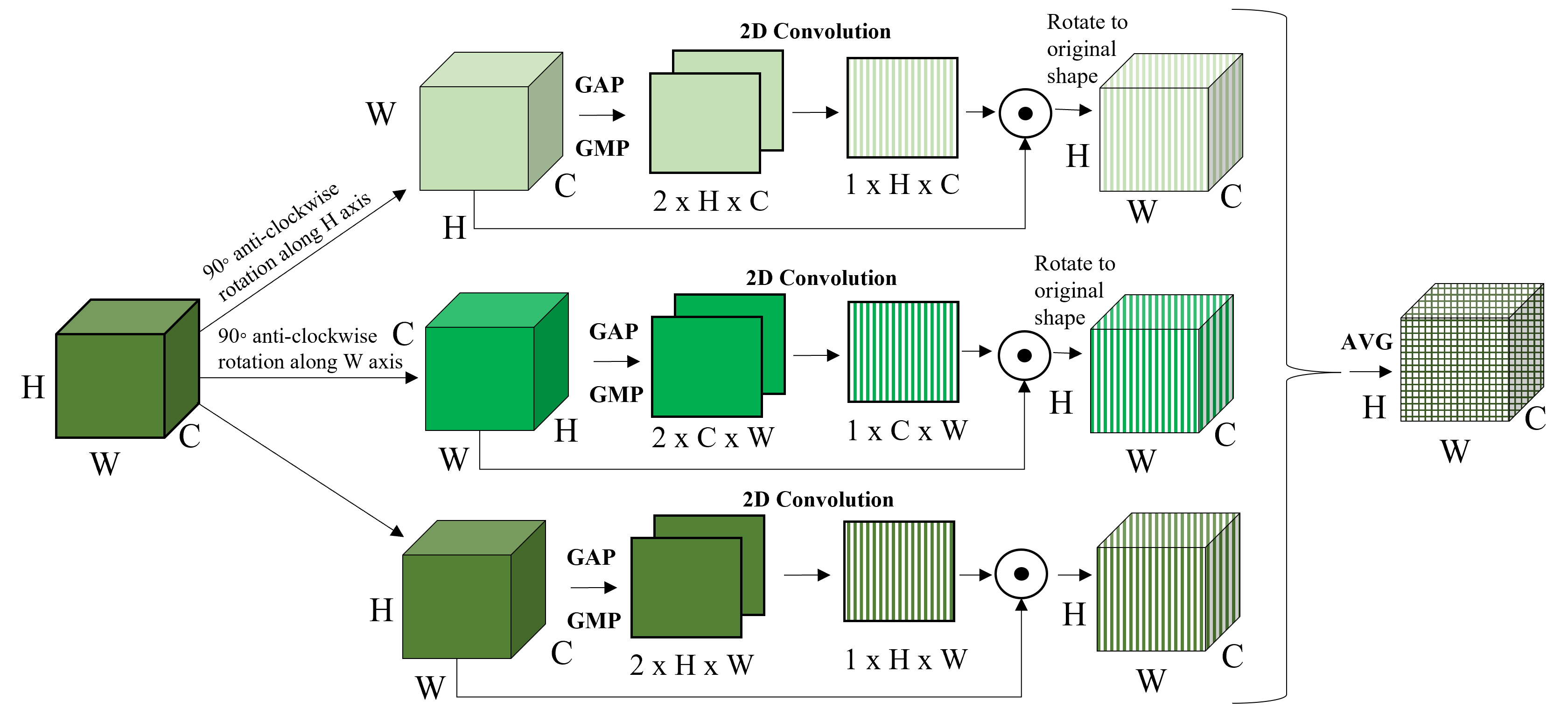}
	\caption{Architecture of the TA attention module, which uses three different branches. The first two branches look simultaneously at the spatial and channel part of the tensor. The third branch looks to the spatial part of the tensor. The weights from the three branches are averaged to output the final tensor. $\odot$ represents element-wise multiplication and GAP stands for Global-Average Pooling and GMP for Global-Max Pooling.}
	\label{fig:ta}
\end{figure}

\subsection{Self-Attention}

Self-attention is a mechanism that relates different positions of a single input to build a different representation of the input. It is an approach originally developed for NLP but with several applications in computer vision. The Transformer architecture  \cite{aiayn} proposed this mechanism with great contextual power. This architecture rapidly became the state-of-the-art in neural network architectures for NLP and a number of authors proposed approaches to to make Transformers usable in computer vision.

The Transformer uses a self-attention mechanism, which relates different positions of the input. Self-attention combines information by using the concepts of Queries, Keys, Values and Multi-Head Self-Attention (MHSA). The multi-head module is simply a concatenation of single attention heads. Each attention head uses self-attention to calculate a score. The scores are then combined to compute the representation in the next layer. Instead of computing with one single attention head, it was found \cite{aiayn} that linearly projecting the Queries, Keys and Values N-ways would benefit the model. This means that there are several self-attention operations being performed at the same time, a characteristic that is useful since the input sequences have multiple relationships and different nuances. With multiple heads, MHSA can model different relations to different parts of the sentence. 

This capacity to relate any arbitrary pair of inputs gives the Transformer architecture the ability to combine any two variables of the previous layers, making it much more flexible that CNNs, where each layer can only process data from a local neighbourhood. However, Transformers do not benefit from the inbuilt bias of CNNs and therefore trade off flexibility for the need to use more data, to offset the lack of biases. The Transformer architecture only processes one-dimensional sequences, so in computer vision this implies that bi-dimensional inputs images need to be flattened into a sequence.

Some recent architectures don't feed the whole down-sampled image to the Transformer and instead divide the image into patches. A linear layer will down-sample those patches, and then feed them individually to the Transformer. AAConv does not make use of patches, whereas newer architectures like ViT, DeiT and ConViT use them.

The Transformer architecture does not have a sense of positioning. Since images are highly-structured data, positional embeddings need to be added to the computations of MHSA layers. Different architectures have different methods of computing these embeddings, and they can be further divided into absolute positional embeddings and relative positional embeddings \cite{ramachandran2019stand}. Absolute positional embeddings encode the position of the patch or pixel in the image, while relative positional embeddings consider the pair-wise distances between the elements in the image, whether it is pixel-level or patch-level \cite{ramachandran2019stand}.

Architectures that make use of self-attention have evolved over time. For convenience, we provide a brief description of the more relevant proposals.

\subsubsection{Attention Augmented Convolutions}

\begin{figure}[htbp!]
	\centering        
	\includegraphics[width=0.3\textwidth]{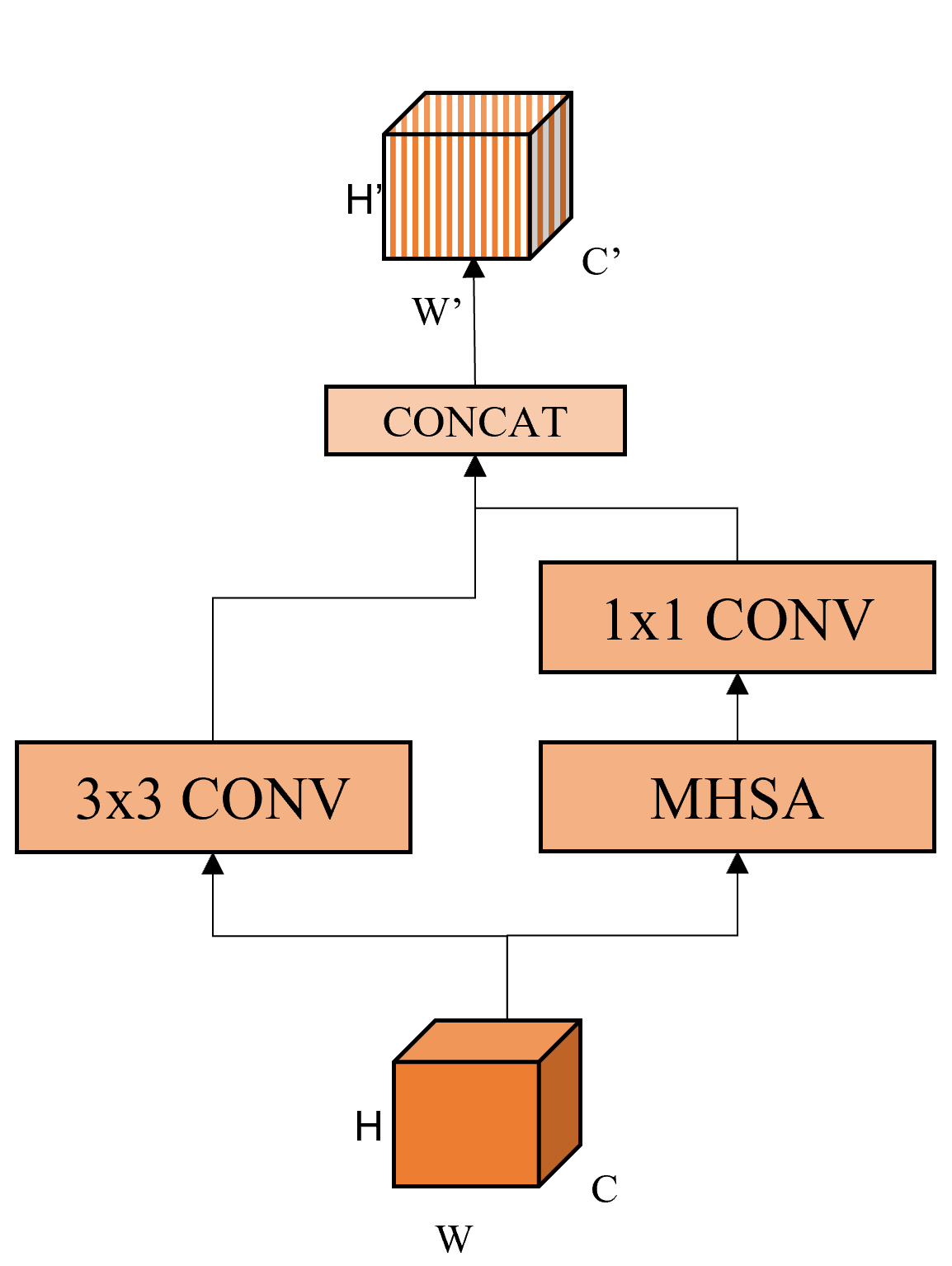}
	\caption{The AA architecture applied to the ResNet. In both the basic and bottleneck block, the $(3\times 3)$ convolutions are augmented with self-attention. MHSA stands for Multi-Head Self-Attention. The dimensions of the output tensor can be different from the input.}
	\label{fig:aa}
\end{figure}

AA \cite{aa} was proposed as an improvement to CNNs and it
is not a new architecture, on itself, but more an augmentation to convolution operations. As such, it will be tested using the ResNet family as the CNN backbone, as was done with the other attention mechanisms described in Section \ref{sec:soft}. In this approach, one or more convolutions of a CNN are replaced with an Attention Augmented Convolution. The new convolution, besides calculating the new tensor with the usual convolution, also processes the input tensor in a MHSA layer. The results from both layers are then concatenated and the resulting tensor goes into the next layer, as shown in Figure \ref{fig:aa}. 

\subsubsection{Vision Transformer}

ViT \cite{vit} was one of the first proposals to build a computer vision model solely based on self-attention. While hybrid architectures like AA would still make use of CNNs as the backbone architecture, ViT shows that good performances can be achieved be stacking multiple transformer-encoder blocks thus rejecting the use of convolutions. 

\begin{figure}[htbp!]
	\centering        
	\includegraphics[width=0.4\textwidth]{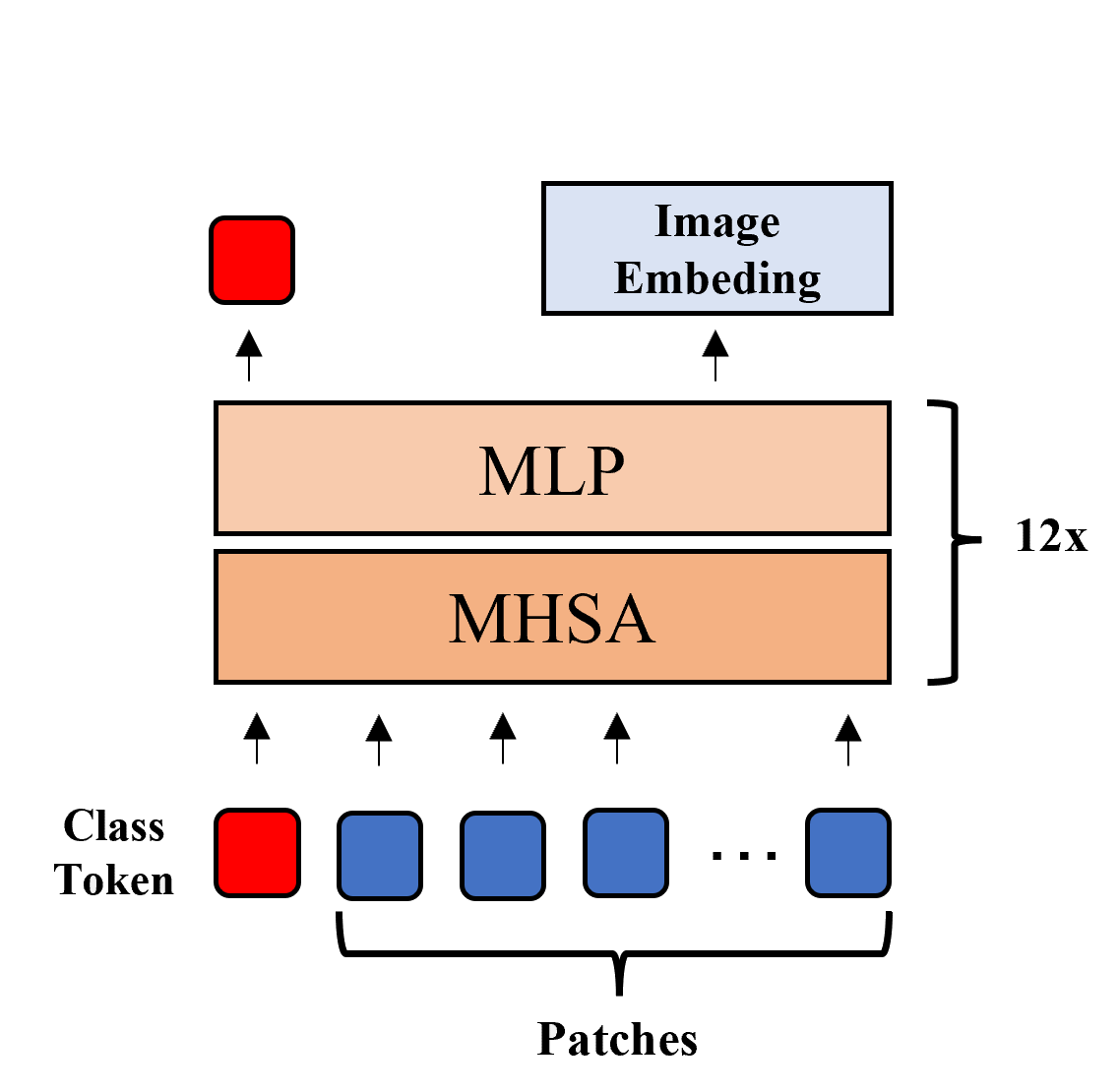}
	\caption{The ViT-Base architecture \cite{convit}. MLP stands for Multi-Layer Perceptron: two linear layers separated by a GeLU activation function. MHSA stands for Multi-Head Self-Attention.}
	\label{fig:vit}
\end{figure}

ViT is composed by 12 transformer-encoder blocks and an MLP head at the end (Figure \ref{fig:vit}). This means that ViT only uses self-attention layers and doesn't use convolutions. However, to be competitive with ResNet on ImageNet-1k with the same number of parameters, ViT needs extensive pretraining.

\subsubsection{Data-Efficient Image Transformer}

\begin{figure}[htbp!]
	\centering        
	\includegraphics[width=0.4\textwidth]{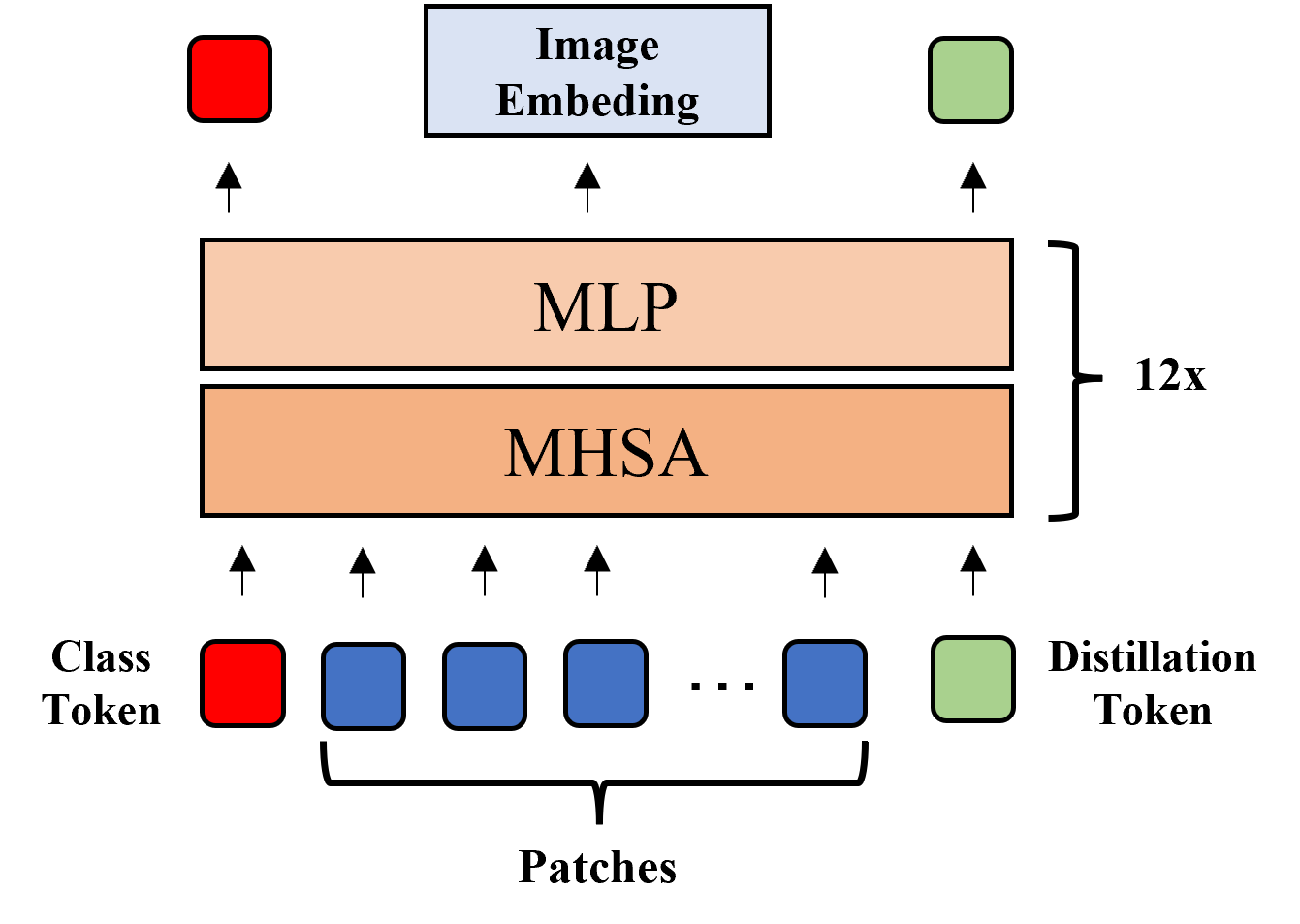}
	\caption{The DeiT-Base architecture \cite{convit}. MLP stands for Multi-Layer Perceptron: two linear layers separated by a GeLU activation function. MHSA stands for Multi-Head Self-Attention.}
	\label{fig:deit}
\end{figure}

DeiT \cite{deit} was proposed as an improvement to ViT. The author  claim that DeiT outperforms CNNs without the need to use ImageNet-21k to pretrain the network. To do so, they use Knowledge Distillation, an approach where a previously trained teacher network (a CNN) transfers its knowledge to the student self-attention based network. The architecture is depicted in Figure \ref{fig:deit}.

\subsubsection{Convolutional Vision Transformer }

\begin{figure}[htbp!]
	\centering       
	\includegraphics[width=0.3\textwidth]{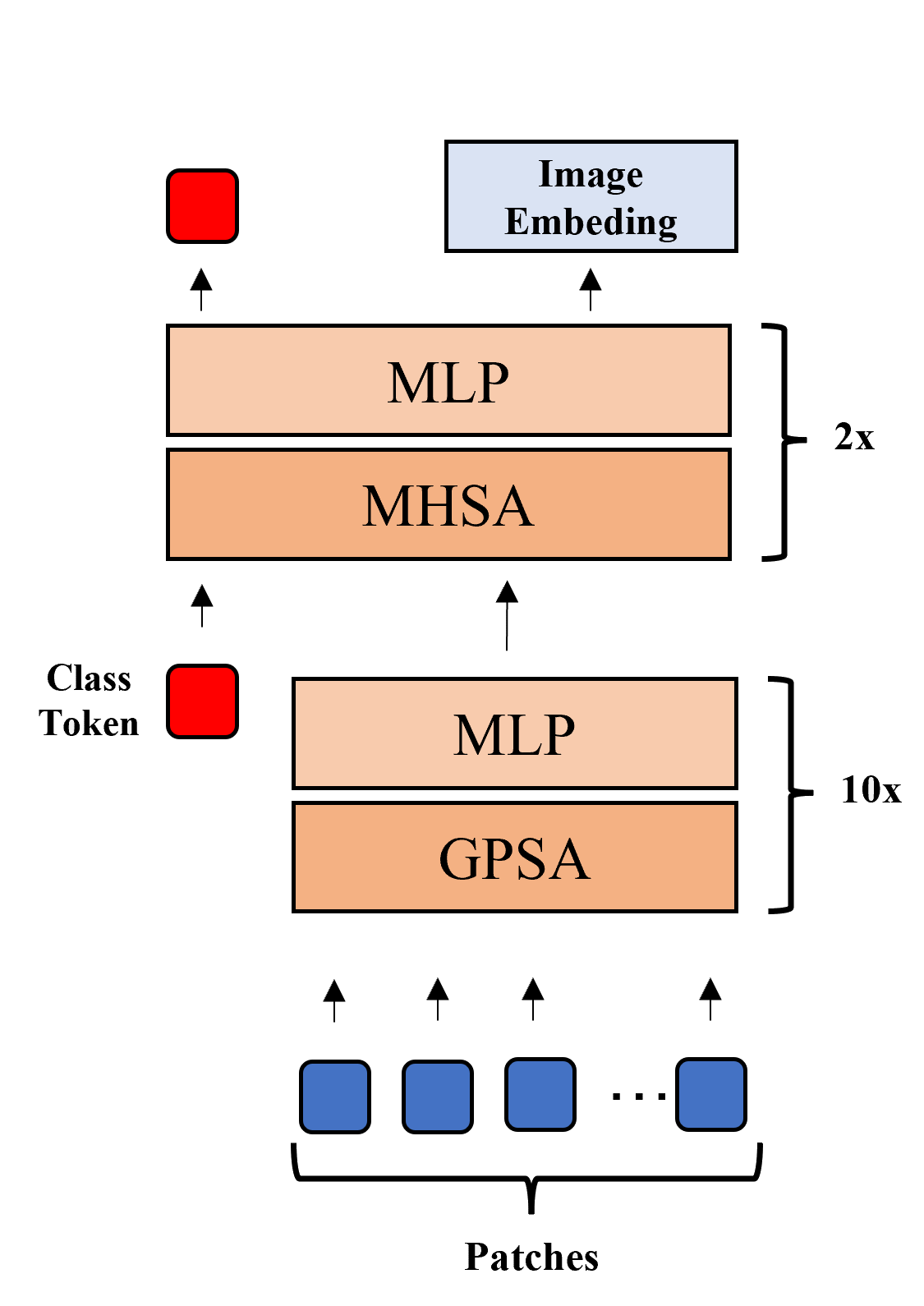}
	\caption{The ConViT-Base architecture \cite{convit}. MLP stands for Multi-Layer Perceptron: two linear layers separated by a GeLU activation function. GPSA is Gated Positional Self-Attention, MHSA stands for Multi-Head Self-Attention.}
	\label{fig:convit}
\end{figure}

ConViT \cite{convit} was also proposed as an improvement to ViT. The idea is to benefit from biases that are inherent to CNNs by letting each layer switch between a local (convolutional) behaviour and a global (attentional) behaviour.
ConViT introduces a new layer named Gated Positional Self-Attention (GPSA) layer. This self-attention layer is able to behave convolutionally, or behave normally like an usual self-attention layer. GPSA layers have a learnable gating parameter which lets the network decide how to behave. This gating parameter is local to each attention head (see Figure \ref{fig:convit}).

\section{Experimental setup}

\subsection{The MNIST Skin Cancer dataset}

Many proposed attention mechanisms have been evaluated and compared using large datasets, such as ImageNet-1k. While this dataset played a fundamental role in Computer Vision, in practice the amount of data available in real-life use cases is rarely the same. Therefore, assessing these architectures in smaller datasets is important in order to evaluate their performance in applications with limited data.

We decided to use The Skin Cancer MNIST (HAM10000) dataset \cite{ham10000} to perform the comparisons.
This dataset represents a good use case to assess the capabilities of attention mechanisms in neural networks,  since the images have a significant amount of noise and, often, the skin lesions are only in a small part of the image. There are also images with hairs and other skin features, which make the classification very challenging. Past research on dermatoscopy focused on distinguishing melanomas, the deadliest type of skin cancer, from benign skin lesions. However, and unlike previous datasets, this one also focuses on non-melanocytic pigmented lesions. Besides identifying malignant from non-malignant, a clinician also needs to make other types of diagnosis, making this dataset more challenging than previous ones.

This dataset contains 10013 training images and was part of the ISIC 2018 classification challenge \cite{isic2018}. The performance of the models was assessed by the challenge server, using a separate test set, whose keys were not available to the competitors. In this challenge, the objective was to classify an image as belonging to one of seven classes: {\em akiec}, {\em bcc}, {\em bkl},  {\em df}, {\em nv}, {\em mel}  and {\em vasc}. More than 95\% of all lesions encountered in clinical practice are present in the classes of this dataset \cite{ham10000}. 

\subsection{Neural network architectures}

The comparisons between the different attention mechanisms were performed using two types of neural network architectures: ResNets and Transformers.

\subsubsection{Residual networks}

The ResNet architecture \cite{resnet} is one of the most popular and effective convolutional neural networks in use. It was first introduced in the ILSVRC 2015 classification challenge and continues to be one of the most used architectures for computer vision tasks. Although many new and more recent architectures have been proposed, ResNet remains as one of the more solid choices for applications in Computer Vision. 

There are many variants of the ResNet architecture. The main differences reside in the type of blocks used and the number of layers. We will use the variants ResNet-34, ResNet-50, ResNet-101 and ResNet-152 (the numbers refer to the number of layers).

The ResNet architecture addresses to problem of the vanishing gradient faced by other CNN architectures when the depth of the networks increases. The vanishing gradient problem makes it difficult to propagate useful gradient information, as the gradient becomes smaller and smaller as it goes throughout the network, making the weights update small or even null. 

To address this problem, the ResNet architecture uses shortcut connections. Shortcut connections decrease the effect of vanishing gradients on deeper networks. They allow gradients to flow directly in a network without passing the convolutional blocks. These connections skip one or more layers, and their outputs are added to the outputs from those stacked layers \cite{resnet}. The connections do not introduce extra parameter computations.

\begin{figure}[H]
	\centering
    \label{fig:basic}	        
    \includegraphics[width=0.2\textwidth]{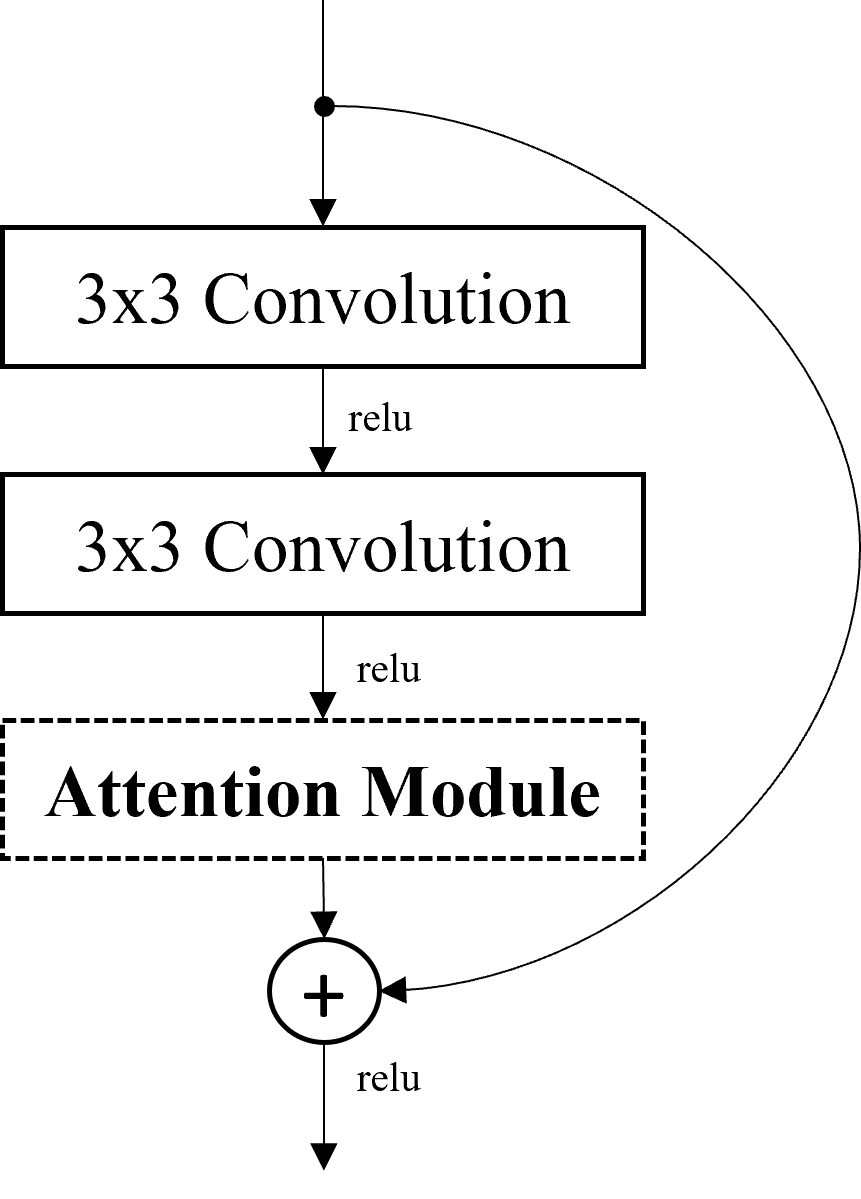}
    \hspace{1cm}
	\includegraphics[width=0.2\textwidth]{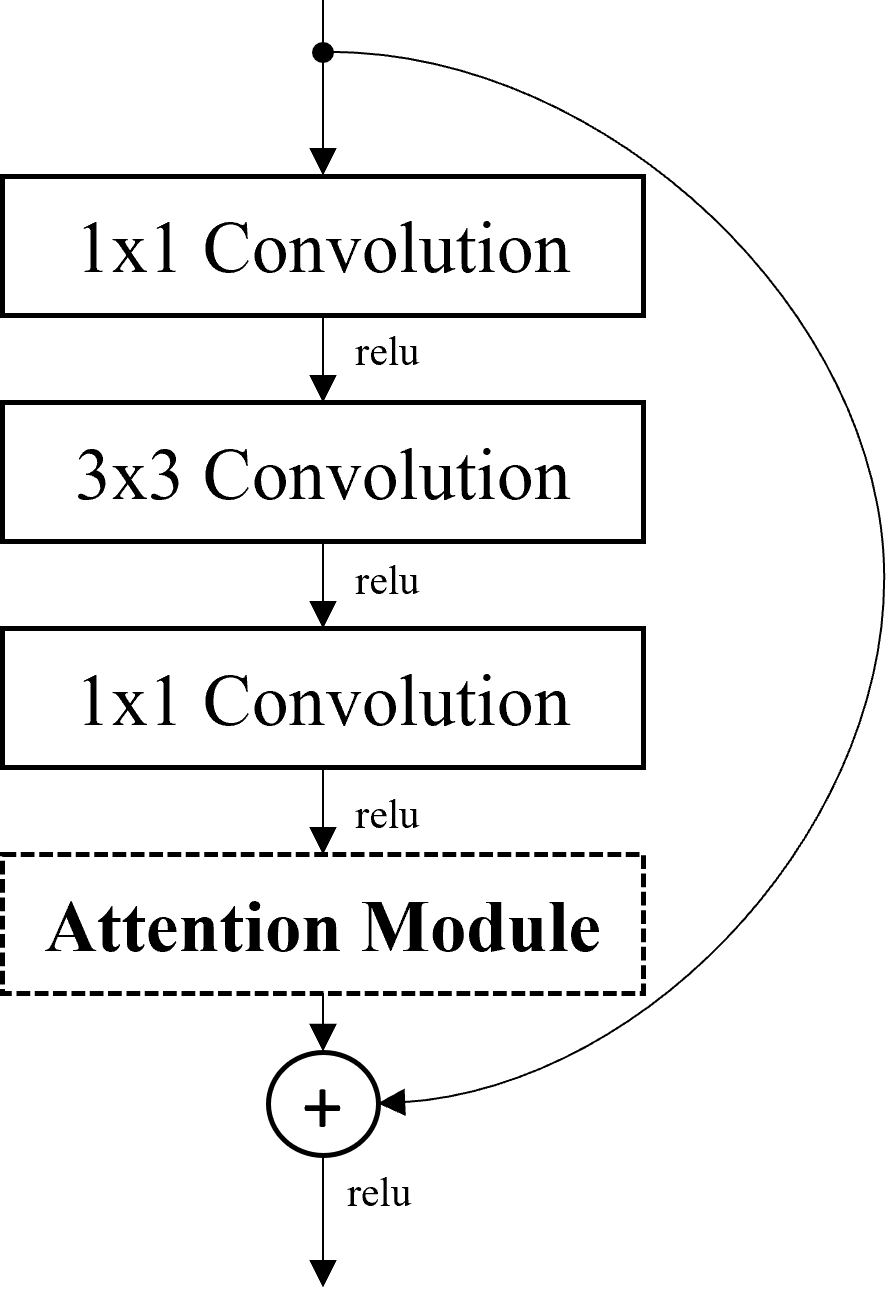}
	\caption{Comparison between Basic (left) and Bottleneck Block (right) of the ResNet architecture.}
	\label{fig:block}
\end{figure}

\Cref{fig:block} shows the basic blocks for the ResNet architecture used. The basic block has a skip connection between two $(3\times 3)$ filter convolutions. For reference, a ResNet-34 is composed of 16 basic blocks, which means 32 layers since each block has two convolutional layers, plus two separate layers, one $(7\times 7)$ convolution at the beginning and a fully connected layer at the end.

Unlike the baseline architectures, deeper ResNets have a bottleneck block. This change is related to concerns in training time. These blocks use one $(3\times 3)$ convolution instead of two. When using bottleneck blocks, a $(3\times 3)$ convolution is inserted between two $(1\times 1)$ convolutions. As \Cref{fig:block} shows, these networks have three layers: $(1\times 1)$, $(3\times 3)$ and $(1\times 1)$. The $(1\times 1)$ layers are responsible for reducing and increasing the dimensions.

One of the most used architectures is the ResNet-50 architecture. This architecture is created by replacing every basic block of the ResNet-34 with a bottleneck block. The ResNet-101 and ResNet-152 are built by increasing the number of bottleneck blocks and therefore making deeper networks with a lot more parameters.

When using ResNets with attention, the attention modules are inserted at the end of each Basic or Bottleneck Block. The attention module is inserted after the two or three convolutions in each block and before the residual connection, as portrayed in \Cref{fig:block}.

\subsubsection{Transformer}

Three different self-attention architectures were tested in this work: 
\begin{itemize}
    \item Vision Transformer (ViT) \cite{vit}. 
    \item Data-Efficient Image Transformer (DeiT) \cite{deit}; 
    \item Convolutional Vision Transformer (ConViT) \cite{convit};
\end{itemize}

\section{Experiments}

In this section, we present the results obtained by the different attention mechanisms in the Skin Cancer MNIST (HAM10000) dataset. We will use NA to stand for 'No Attention'. Therefore ResNet + NA refers to the baseline, the plain original ResNet without any augmentations or attention modules

\subsection{Pre-processing and parameterization}

All the images were resized to $224\times 224$ resolution and the following transformations were used in the training set: RandomHorizontalFlip(), RandomVerticalFlip(), and RandomRotation(20). The pixel mean and standard deviation of the images were also used to normalize the dataset.

For the ResNet family and all the attention mechanisms variants, the learning rate chosen was $r = 0.1$. This value corresponds to the most common learning rate for training ResNet models and is also the standard choice used in previous work that uses attention modules. As other authors have done, we applied a learning rate policy that decays the base learning rate every 30 epochs for a total of 100 epochs. Due to computational restrictions, the models were trained with a smaller batch size of 32. Following previous work \cite{cbam,ta,squeeze,eca}, the model was trained with the SGD optimizer, momentum $0.9$ and weight decay of $0.0001$. When ImageNet-1k \cite{imagenet} was used to pre-train the models, instead of using $r = 0.1$, a smaller learning rate was used. Three different learning rates were tested (0.01, 0.001, 0.0001) and the one that obtained the best performance was $r = 0.001$. The learning rate decay policy was kept the same while fine-tuning.

For the self-attention based models \cite{vit,convit}, the hyper-parameters used were different. To train the models in ImageNet-1k, we used learning rates of $3 \times 10^{-3}$ in ViT, and $5 \times 10^{-4} \times \frac{b}{512}$, where $b$ is the batch size, in DeiT and ConViT. With a batch size of 32, this leads to a learning rate of $3 \times 10^{-5}$ for both DeiT and ConViT for training with ImageNet-1k. 

Following the reasoning employed in the ResNet models, the self-attention weights pre-trained on ImageNet-1k were fine-tuned with a smaller learning rate. Three different learning rates were tested ($3 \times 10^{-5}$, $3 \times 10^{-6}$, $3 \times 10^{-7}$). For the ViT variants, the learning rate used was $3 \times 10^{-6}$. For DeiT and ConViT, the learning rate used was $3 \times 10^{-7}$. No learning rate decay was used to fine-tune the weights. 

The models were trained with the AdamW optimizer with a batch size of 32 and weight decay of 0 for ViT and 0.05 for DeiT and ConViT. The choice of optimizer and weight decay was taken from the articles that proposed these architectures. While the authors of ViT clearly state that a weight decay of 0 was used while fine-tuning, we did not find that information for ConViT and DeiT, so the weight decay used for the training on ImageNet-1k was replicated.

\Cref{tab:random_skin} shows the results from the models initialized with random weights. Three variants of the ResNet family were tested, with and without the attention modules added.

\subsection{Results obtained of the ResNet architecture with attention mechanisms \label{sec:att_results}}

For the first two ResNet variants, ResNet-34 and ResNet-50, the attention mechanisms consistently improved the performance of the networks on the test set. Regarding the ResNet-34 variant, the ResNet-34 + SE model was the best performing variant. The SE augmented model had a balanced accuracy of 0.539, 0.073 more than the baseline. For the ResNet-50 variant, ResNet-50 + CBAM model had a balanced accuracy of $0.566$, $0.044$ more than the baseline.

However, the ResNet-101 variant exhibited a different behavior. The improvements brought by the attention mechanisms were less consistent and the only models that outperformed the baseline were CBAM and AAConv. The AAConv variant gave a surprising performance boost to the original network. ResNet-101 + AAConv had a balanced accuracy of $0.622$, $0.074$ more than the baseline. This increase in performance could be related to the increase in the number of parameters. The AAConv architecture has a parameter overhead of 22.94 million parameters, when compared to the plain ResNet-101 + NA.

\begin{table}[htbp]
\centering
\caption{Scores in three variants of the ResNet family with different attention modules and random weights initialization. The best result is in bold.}
\label{tab:random_skin}
\def\arraystretch{1.1}
\begin{tabular}{|c|c|c|c|}
\hline
\rowcolor[HTML]{ECF4FF} 
\multicolumn{1}{|l|}{\cellcolor[HTML]{ECF4FF}Architecture} & Modules & Parameters & Balanced Accuracy  \\ \hline \hline
                                                           & NA                & 21.28M     & 0.466          \\ 
                                                           & SE                & 21.45M     & \textbf{0.539} \\ 
                                                           & CBAM              & 21.45M     & 0.514          \\ 
                                                           & ECA               & 21.28M     & 0.509          \\ 
                                                            & TA                & 21.29M     & 0.518          \\  
\multirow{-5}{*}{ResNet-34}                                 & AA                & 40.451M     & 0.469
\\ \hline
                                                           & NA                & 23.52M     & 0.522          \\ 
                                                           & SE                & 26.05M     & 0.553          \\ 
                                                           & CBAM              & 26.05M     & \textbf{0.566} \\ 
                                                           & ECA               & 23.52M     & 0.546          \\ 
                                                           & TA                & 23.52M     & 0.562          \\ 
\multirow{-6}{*}{ResNet-50}                                & AA                & 33.87M     & 0.544          \\ \hline
                                                           & NA                & 42.50M     & 0.548          \\ 
                                                           & SE                & 47.29M     & 0.538          \\ 
                                                           & CBAM              & 47.29M     & 0.550          \\ 
                                                           & ECA               & 42.51M     & 0.536          \\ 
                                                           & TA                & 42.52M     & 0.535          \\ 
\multirow{-6}{*}{ResNet-101}                               & AA                & 65.44M     & \textbf{0.622} \\ \hline
\end{tabular}
\end{table}

\Cref{tab:imagenet_skin} shows the results from the models initialized with pre-training on ImageNet-1k. Contrary to what was done with the initial random weights, only the ResNet-50 was used in this experiment, since most attention models proposed do not provide the weights for the pre-trained versions in every ResNet variant. Besides ResNet-50 being one of the most popular and common variants of the ResNet family, it was the one for which most attention modules had available the weights for the network pre-trained on ImageNet-1k. To increase the significance of the results, three different runs were performed for each attention mechanism. \Cref{tab:imagenet_skin} shows the mean of the balanced accuracy and the corresponding standard deviation from three different training runs.


With models pre-trained on ImageNet, the performance increase given by the attention mechanisms is not consistent, as only the ECA attention module provided a modest improvement in the performance. Most often, the attention mechanisms degrade the performance when compared to the baseline. 

\begin{table}[htbp]
\centering
\caption{Scores from three different training sessions in ResNet-50 with different attention mechanisms pre-trained on ImageNet-1k. The Balanced Accuracy score is followed by the standard deviation. The best result is in bold.}
\label{tab:imagenet_skin}
\def\arraystretch{1.1}
\begin{tabular}{|c|c|c|c|}
\hline
\rowcolor[HTML]{ECF4FF} 
\multicolumn{1}{|l|}{\cellcolor[HTML]{ECF4FF}Architecture} & Modules & Parameters & B.Accuracy  \\ \hline \hline
                                                           & NA                & 23.52M     & 0.647 $\pm$ 0.022          \\ 
                                                           & SE                & 26.05M     & 0.622 $\pm$ 0.018          \\ 
                                                           & CBAM              & 26.05M     & 0.595 $\pm$ 0.021          \\ 
                                                           & ECA               & 23.52M     & \textbf{0.662} $\pm$ 0.015          \\ 
                                                           & TA                & 23.52M     & 0.644 $\pm$ 0.038 \\ 
\multirow{-6}{*}{ResNet-50}                                & AA                & 33.87M     & 0.626 $\pm$ 0.026          \\ \hline
\end{tabular}
\end{table}

\begin{figure}[htbp!]
	\centering        
	\includegraphics[width=0.45\textwidth]{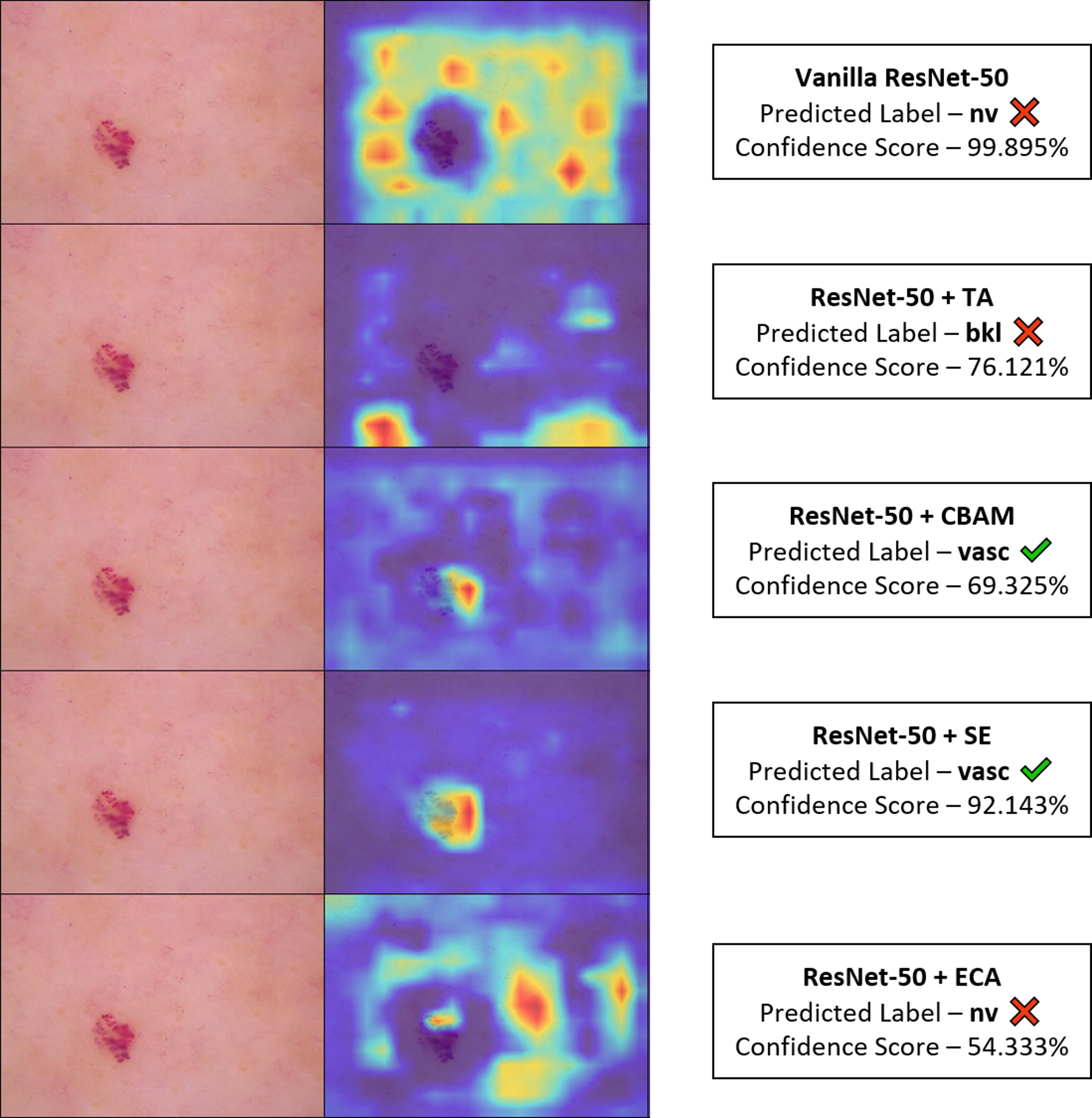}
	\caption{GradCAM visualization of the validation set in the VASC class.}
	\label{fig:gcam_vasc}
\end{figure}

One of the objectives of this work was to verify if attention mechanisms change in a visible way the weight the models give to different parts of the input. Therefore, we used the GradCAM technique to analyze the weight given to different parts of the input, by computing the attention maps.  \Cref{fig:gcam_vasc} depicts the GradCAM result, showing the parts of the input that have more weight on the decision. In this case, a ResNet-50 with random weights initialization was used. To make the attention maps more detailed, they were generated with the images at their original resolution. AA is not used with GradCAM since the input size had to be fixed to the original $224\times 224$ resolution. Many other results were obtained for different samples of different classes. In general it seems that the attention modules do in fact focus the attention of the network on more specific places within the image.

\subsection{Results obtained with self-attention mechanisms}

\Cref{tab:architecture_skin} compares the results on different variants of the ResNet architecture and the self-attention models, ViT, DeiT and ConViT. All the variants were pre-trained on ImageNet-1k. The weights come from the work of Ross Wightman \cite{rw2019timm}. In these results, it is clear that increasing the number of parameters does translate to an increment in the performance of the models. The model that held the best performance was ViT.

Although the proponents of ViT stated that pre-training on ImageNet-21k was need to surpass the performance of ResNets, our results do not confirm this result. \Cref{tab:1kvs21k} shows the results of the models pre-trained on ImageNet-21k compared against the ones pre-trained on ImageNet-1k. This comparison was performed to verify if there were any benefits from pre-training in a much larger dataset. These results show that there seems to be no direct benefit from using ImageNet-21k in the HAM10000 dataset. Most often, the pre-training on this dataset degrades the performance.

\begin{table}[htbp]
\centering
\caption{Scores in different architectures with pre-training on ImageNet-1k. The best result is in bold.}
\label{tab:architecture_skin}
\def\arraystretch{1.1}
\begin{tabular}{|c|c|c|}
\hline
\rowcolor[HTML]{ECF4FF} 
Architecture & Parameters & B.Accuracy  \\ \hline \hline
ResNet-34    & 21.28M     & 0.603       \\ 
ResNet-50    & 23.52M     & 0.646       \\ 
ResNet-101   & 42.50M     & 0.675       \\ 
ResNet-152   & 58.15M     & 0.682       \\ \hline
ViT-Tiny     & 5.53M      & 0.651       \\ 
ViT-Small    & 21.67M     & 0.677       \\ 
ViT-Base     & 85.80M     & \textbf{0.737} \\ 
ViT-Large    & 303.31M    & 0.703       \\ \hline
DeiT-Tiny    & 5.53M      & 0.621       \\ 
DeiT-Small   & 21.67M     & 0.658       \\ 
DeiT-Base    & 85.80M     & 0.687       \\ \hline
ConViT-Tiny  & 5.52M      & 0.637       \\ 
ConViT-Small & 27.35M     & 0.683       \\ 
ConViT-Base  & 85.77M     & 0.699       \\ \hline

\end{tabular}
\end{table}

\begin{table}[htbp]
\centering
\caption{Scores in ViT with pre-training on ImageNet-1k and ImageNet-21k. The best results are in bold.}
\label{tab:1kvs21k}
\def\arraystretch{1.1}
\begin{tabular}{|c|c|c|}
\hline
\rowcolor[HTML]{ECF4FF} 
Architecture & ImageNet-1k &  ImageNet-21k  \\ \hline \hline
ViT-Tiny     & 0.651          &  \textbf{0.660}     \\ 
ViT-Small    & \textbf{0.677}          &  0.671     \\ 
ViT-Base     & \textbf{0.737} &  0.696    \\ 
ViT-Large    & \textbf{0.703}          &  0.680     \\ \hline

\end{tabular}
\end{table}

\subsection{Analysis of results}

The results obtained in this section have shown that ViT-Tiny and ConViT-Small stand out as models with low parameter complexity that exhibit good performance on this dataset. ViT-Tiny reaches higher accuracy while having fewer parameters than ResNet-34 and ResNet-50. The same is true for ConViT-Small,  which exhibits higher accuracy than ResNet-101 with almost half the parameters.

These results show that the ViT family of architectures, based on self-attention, outperforms all other architectures in this problem. 

\section{Conclusions and Future Work}

We have presented an assessment of the effectiveness of attention and self-attention mechanisms in the HAM10000 dataset.

Overall, the results from the attention modules are not as consistent as expected. When using networks pretrained on ImageNet-1k, most of the architectures augmented with attention modules under-perform the basic model or result in modest improvements. However, in smaller networks, initialized with random weights, the performance increase is noticeable and consistent for all attention modules. The behaviour in these use cases is also verified using GradCAM \cite{gradcam}, which shows that the attention maps on the HAM10000 dataset seem to be more relevant than the ones that correspond to plain networks. The results suggest that the attention modules reduce the noise in the images and look at more specific regions and at specific shapes within the images. Unfortunately, the improvements are not consistent across the board in the different variants of the ResNet architecture.


TA and ECA were proposed as enhancements to older attention modules but the results obtained in this work show no significant improvements over their older counterparts, in this particular dataset.

Another difficulty with the use of attention modules is related to the fact that several of them do not have pre-trained weights for all the CNN architectures and variants. Often the researcher will need to train the network together with the attention module in a large dataset, such as ImageNet, a not very attractive proposition, as training on ImageNet is computationally expensive.



Self-attention mechanisms, on the other hand, seem much more promising. Many different architectures built around the ViT are being published. However, these architectures also have sample efficiency problems, as they do not have inherent topology-related biases, which help in learning with little data \cite{vit,convit}. In the HAM10000 dataset, ViT exhibited the best performance among all the tested models. The good results from this architecture could be related to the fact that self-attention, unlike convolutions, looks more at the shape of the image than at the texture \cite{object_shape}. The form of the skin lesions is essential to differentiate between different classes. Surprisingly, many of the self-attention architectures surpassed the performance of ResNet based architecture, with significantly less parameters. 



The results obtained in this work suggest that the future developments in use of attention in computer vision-related problems should point towards self-attention. These are the architectures with the most promising results. In particular, the good performance of ViT in the HAM10000 dataset shows significant potential. 

\bibliographystyle{IEEEtran}
\bibliography{bibfile}

\end{document}